\documentclass{styles/svproc}
\usepackage{url}

\usepackage{algorithm2e}
\usepackage{graphicx}
\usepackage{amsmath}
\usepackage{hyperref}

\begin{document}
\mainmatter              
\title{Case-Based Inverse Reinforcement Learning Using Temporal Coherence}
\titlerunning{Case-Based Inverse Reinforcement Learning Using Temporal Coherence}

\author{
Jonas Nüßlein \footnote{Corresponding author} \and
Steffen Illium\and
Robert Müller\and
Thomas Gabor\and
\newline
Claudia Linnhoff-Popien
}

\authorrunning{Nüßlein et al.} 
%
\tocauthor{Jonas Nüßlein, Steffen Illium, Robert Müller, Thomas Gabor and Claudia Linnhoff-Popien}
\institute{LMU Munich, Institute of Computer Science, Germany,\\
\email{\{jonas.nuesslein, steffen.illium, robert.mueller, thomas.gabor, linnhoff\}@ifi.lmu.de}}

\maketitle              

\begin{abstract}
Providing expert trajectories in the context of Imitation \linebreak Learning is often expensive and time-consuming. The goal must therefore be to create algorithms which require as little expert data as possible. In this paper we present an algorithm that imitates the higher-level strategy of the expert rather than just imitating the expert on action level, which we hypothesize requires less expert data and makes training more stable. As a prior, we assume that the higher-level strategy is to reach an unknown target state area, which we hypothesize is a valid prior for many domains in Reinforcement Learning. The target state area is unknown, but since the expert has demonstrated how to reach it, the agent tries to reach states similar to the expert. Building on the idea of Temporal Coherence, our algorithm trains a neural network to predict whether two states are similar, in the sense that they may occur close in time. During inference, the agent compares its current state with expert states from a Case Base for similarity. The results show that our approach can still learn a near-optimal policy in settings with very little expert data, where algorithms that try to imitate the expert at the action level can no longer do so.

\end{abstract}

\section{Introduction}

In Reinforcement Learning (RL), the goal of the agent, given a Markov Decision Process, is to maximize the expected cumulative reward. The higher the expected reward, the better the agent's policy. In Imitation Learning, on the other hand, the agent does not have access to a reward signal from the environment. Instead, it either has access to an expert who can be asked online for the best action for a given state or a set of trajectories generated by the expert is available. Imitation Learning has been proven to be particularly successful in domains where the demonstration by an expert is easier than the construction of a suitable reward function \cite{arora2021survey}. There are two main approaches to Imitation Learning: Behavioral Cloning (BC) \cite{pomerleau1991efficient} and Inverse Reinforcement Learning (IRL) \cite{fu2017learning,arora2021survey,abbeel2004apprenticeship}. In BC, the agent learns via supervised learning to produce the same actions that the expert would have produced. The advantage of this approach is that no further rollouts in the environment are necessary. However, the approach suffers greatly from compounding error, i.e., the slow drift of states visited by the expert \cite{ross2011reduction}. In the second approach, IRL, a reward function is learned under which the expert is uniquely optimal. Then, a policy can be learned using classical Reinforcement Learning and this reconstructed reward function. However, the drawback of this approach is that it usually requires a lot of rollouts in the environment, as it often includes RL as a subroutine.

GAIL \cite{ho2016generative} is another approach to Imitation Learning. It builds on the ideas of Generative Adversarial Networks. In this approach, a policy and a discriminator are learned. The goal of the discriminator is to be able to distinguish state-action pairs of the expert from state-action pairs of the agent, while the goal of the policy is to fool the discriminator. GAIL requires expert actions, but there is an extension, named GAIfO, which does not \cite{torabi2018generative}. While GAIL discriminates between state-action pairs produced by the agent or the expert respectively, GAIfO does so with state transitions. In this paper we consider, as GAIfO does, the setting where no expert actions are available to the agent. This setting is also called Learning from Observation (LfO) or Imitation from Observation (IfO) \cite{yang2019imitation,torabi2018generative}.

\ \\
Providing expert trajectories is often very expensive and time-consuming, especially if the expert is a human. The goal must therefore be to create algorithms which require as little expert data as possible.

The aim of this paper is to present an algorithm that imitates the higher-level strategy of the expert rather then just imitating the expert on action level.

Our motivation for this is that we hypothesize that it takes less expert data to learn the higher-level strategy than to imitate the expert on action level. We also hypothesize that it makes the training more stable, with less ``forgetting'' of what has already been learned. As a prior for the higher-level strategy, we assume that the higher-level strategy is to reach an unknown target state area, which we hypothesize is a valid prior for many domains in Reinforcement Learning.

We present an algorithm that learns these higher-level strategies from expert trajectories. To prove that the algorithm does not imitate the expert on action level, we consider a special setting of Imitation Learning, which is characterized by incomplete expert trajectories. Here, the agent does not see every state of the expert trajectory, but, for example, only every fifth. Thus, it cannot imitate the expert on action level.

\ \\
The idea behind machine learning is to derive general rules from a large amount of data, which can then be applied to new, unknown scenarios. This induction-based learning principle differs from Case-Based problem solving. In Case-Based Reasoning, a set of problems solved in the past is stored in a database. If a new, unknown problem is to be solved, the problem most similar to the current situation is retrieved from the database and used to solve the current problem. Applications of Case-Based Reasoning range from explaining neural network outputs \cite{li2018deep,keane2019case} over financial risk detection \cite{li2021data} to medical diagnosis \cite{choudhury2016survey}. In our algorithm we build on ideas from Case-Based Reasoning as well as on the idea of Temporal Coherence.

Temporal Coherence \cite{goroshin2015unsupervised,mobahi2009deep,zou2011unsupervised} originates from Video Representation Learning, where the idea is that two images, which occur shortly after each other in a video, are very likely to show the same object or objects. The two images should therefore have a similar representation. On the other hand, distant images should have different representations. The combination of this convergence and divergence, also called contrastive learning, can be used as a self-supervised training signal to learn semantically meaningful representations \cite{knights2021temporally}.
\ \\

Our contribution in this paper is twofold. First, we propose the setting with incomplete expert trajectories without expert actions as a way to prove that the agent really learns the expert's strategy and does not imitate the expert on action level. The prior we are using for the higher-level strategy is to reach an unknown target state area. Second, we present an algorithm that can learn such higher-level strategies and we test it on four typical domains of RL. The results show that our approach can still learn a near-optimal policy in settings with very little expert data, where IRL algorithms that try to imitate the expert at the action level can no longer do so.

\section{Background}

In this section we want to provide a brief introduction to Markov Decision Processes (MDP) \cite{arulkumaran2017brief}. A MDP is a tuple $(S,A,T,R,\gamma)$. $S$ is a set of states, combined with a distribution of starting states $p(s_0)$. $A$ is a set of actions the agent can perform. $T$ is the transition function of the environment which computes the next state $s_{t+1}$ given a state $s_t$ at time $t$ and an action $a_t$: $T(s_{t+1}|s_t,a_t)$. The property of $T$ that the computation of $s_{t+1}$ depends only on the last state $s_t$ and not on $s_{\tau < t}$ is also called the Markov property, hence the name Markov Decision Process. $r_t=R(s_t,a_t,s_{t+1})$ is a reward function and $\gamma \in [0;1]$ is a discount factor. If $\gamma < 1$, immediate rewards are preferred compared to later rewards. An agent acts in a MDP using its policy $\pi$. The policy is a function which outputs an action $a$ given a state $s$: $\pi(s)=a$. MDPs are often episodic, which means that the agent acts for $T$ steps, after which the environment is reset to a starting state. The goal of the agent in a MDP is to maximize the expected return by finding the policy
\begin{equation}
    {\pi}^* = \underset{\pi}{\mathrm{argmax}}\ E[R|\pi]
\end{equation}
where the return R is calculated via:
\begin{equation}
    R=\sum_{t=0}^{T-1}{\gamma^tr_{t+1}}
\end{equation}

\section{Related Work}

\textbf{Combination of Case-Based Reasoning (CBR) and Reinforcement \linebreak Learning (RL)}: In \cite{bianchi2009improving} the authors use Case-Based Reasoning (CBR) in the setting of Heuristic Accelerated Reinforcement Learning, where a heuristic function assists in action selection to accelerate exploration in the state-action space. 
In \cite{auslander2008recognizing}, Case-Based Reasoning is used to efficiently switch between previously stored policies learned with classical RL. A similar approach is taken by \cite{wender2014combining}.
Most Imitation Learning algorithms try to imitate the expert skill step-by-step. In \cite{lee2019follow}, a hierarchical algorithm is presented where this goal is mitigated. Instead, a policy is learned that reaches sub-goals, which in turn are sampled by a meta-policy from the expert demonstrations.
\ \\

\textbf{Temporal Coherence in Reinforcement Learning}: Some papers have already investigated the use of Temporal Coherence in the context of Reinforcement Learning. For example, in \cite{florensa2019self} it was proposed to learn an embedding for the inputs of the Markov Decision Process, such that the euclidean distance in the embedding space is proportional to the number of actions the current agent needs to get from one state to the other. The byproduct of this is a policy that can theoretically reach any previously seen state on demand. A similar idea is followed in the context of goal-conditioned RL: In \cite{lee2021generalizable} a proximity function $f(s,g)$ is learned that outputs a scalar proportional to the distance of the state $s$ to the goal $g$. The distance then serves as a dense reward signal for a classical RL agent. This is especially beneficial when the environment's reward function is sparse.

In \cite{sermanet2018time}, a special setting is considered where multiple observations are available simultaneously, showing the same state from different perspectives. An embedding is then learned so that contemporaneous observations have the same embedding and temporally distant observations have different embeddings. Thus, a perspective-invariant representation is learned, which contains semantic information. That paper also considers the case where only one perspective is available. In this case, the embeddings of two nearby inputs should be as similar as possible and temporally distant inputs should be as dissimilar as possible. We build on this idea of Temporal Coherence, although we do not learn an embedding. \cite{dwibedir2018self} extends the idea of \cite{sermanet2018time} to input sequences to contrast movements.

In \cite{savinov2018episodic}, the concept of Reachability Networks is already introduced, i.e., a network that classifies whether two states can occur in short succession in a trajectory. This network is then used as a curiosity signal to guide exploration in sparse reward domains. We build on this concept, but use it differently. While in \cite{savinov2018episodic} the agent searches for dissimilar states, the goal of the agent in our approach is to reach similar states (compared to expert states).
\ \\

\textbf{Curriculum via Expert Demonstrations}: As we will see in the next section, the reconstructed reward function in our approach can be interpreted as an implicit curriculum. A related approach, which creates an explicit curriculum using expert demonstrations, is \cite{dai2021automatic}. In that paper the expert trajectory is divided into several sections and state resetting to expert states is used to increase the difficulty of reaching the goal state. The sector from which expert states are sampled for resetting is gradually pushed away from the goal as the curriculum progresses. A similar approach is \cite{hermann2020adaptive}, which again uses resetting to starting states of varying difficulty.
\ \\

\textbf{Unsupervised Perceptual Rewards for Imitation Learning}: the closest work to ours is \cite{sermanet2016unsupervised}. In that paper the authors examine how to use pre-trained vision models to reconstruct a reward function from few human video demonstrations. They do so by first splitting the human demo videos into segments, then selecting features of a pre-trained model which best discriminate between the segments and then using a reward function, which is based on these selected features, to learn a policy via standard RL algorithms. The biggest difference to our algorithm is that \cite{sermanet2016unsupervised} reconstructs the reward function entirely before training in the RL domain. In contrast, we learn the reward function and the policy at the same time.

\section{Case-Based Inverse Reinforcement Learning (CB-IRL)}

In this work, we consider a special setting of Imitation Learning that is characterized by two main features.
First, there are no expert actions available to the agent and, second, the expert trajectories are incomplete, i.e., from the original sequence of MDP states of the expert $[s_0, s_1, s_2, ..., s_T]$, the agent only sees, for example, every fifth state: $[s_0, s_5, s_{10}, ...]$. This makes it impossible for the agent to imitate the expert at the action level. Given such a setting, we now propose the algorithm Case-Based Inverse Reinforcement Learning (CB-IRL). The architecture of CB-IRL consists of the Case Base ($C$) and two neural networks, the Equality Net ($E$) and the Policy ($\pi$), see Figure 1. $C$ is filled with the expert trajectories.
\ \\

The basic idea is that the agent should not act in every step exactly as the expert would do, but instead imitate the higher-level strategy of the expert. We chose the task of reaching a target state area as the prior for the higher-level strategy. For example, for the OpenAI domain  `MountainCar' the target state area are the states where the car is on top of the mountain. For the Atari game `Pong' the target state area would be the states where the agent has 21 points. The agent does not know the target state area, but since the expert has demonstrated how to reach the target state area, CB-IRL trains the agent to reach similar states as the expert.

Two states are ``similar'' in the context of Reinforcement Learning if it takes only few steps (actions) to get from one state to the other. Other approaches \cite{florensa2019self,sermanet2018time,dwibedir2018self} try to learn a state-embedding such that the euclidean distance of the representations is proportional to the number of steps needed to get from one state to the other. We take a different approach and instead train a neural network that accepts two states $s_1$ and $s_2$ as input and outputs a scalar $E(s_1,s_2) = d$ ; $E:S\times S \rightarrow [0;1]$ to classify whether $s_2$ can be reached within $\textit{windowFrame}$ steps from $s_1$. Thus, this is a classification and not a regression. We believe a classification is easier and more stable to learn compared to a regression, since it suffers less from the ``moving target'' problem. For example if we would predict the number of steps which are required to go from one state to the other, the target of this supervised learning tasks is heavily based on the current performance of the agent. In contrast, for near/far classification, it does not matter if the states are, for example, 30 or 40 steps apart if $\textit{windowFrame}=10$. In both cases the state pair gets the target 0 for supervised learning, since it shall be classified as dissimilar.

\SetKwComment{Comment}{/* }{ */}
\RestyleAlgo{ruled}
\begin{algorithm}
\caption{CB-IRL}\label{alg:two}
\KwData{Case-Base $C$ (containing expert trajectories)}
\KwResult{Policy $\pi$, Equality Net $E$}
\While{training}{
    $s \gets$ sample start state\\
    $r_{pre} \gets $Reward$(s)$\\
    $trajectory \gets [s]$\\
    \While{episode is not finished}{
        $a \gets \pi(s)$\\
        $s' \gets$ execute $a$\\
        $trajectory.append(s')$\\
        $r_{post} \gets $Reward$(s')$\\
        $r \gets r_{post} - \alpha * r_{pre}$\\
        use $(s,a,r,s')$ for training $\pi$\\
        $s \gets s'$\\
        $r_{pre} \gets r_{post}$\\
    }
    append $trajectory$ to the Replay Buffer of $E$\\
    train $E$ using the Replay Buffer, $C$ and the hyperparameters $windowFrame$ and $\nu$\\
}
\ \\

\DontPrintSemicolon
\SetKwProg{Fn}{Function}{:}{}
\SetKwFunction{FMain}{Reward}
\Fn{\FMain{$s$}}{
    $mostSimilar = \mu$\;
    $similarity = \tau$\;
    \ForEach{trajectory $\in C$}{%
        \ForEach{$o_e^{(i)} \in $ trajectory}{
            \If{$E(s, o_e^{(i)}) > similarity$}{
                $mostSimilar = i$\;
                $similarity = E(s, o_e^{(i)})$\;
            }
        }
    }
    \KwRet $mostSimilar$\;
}
\end{algorithm}

\begin{figure}[bt]
\centering
\includegraphics[scale=0.37]{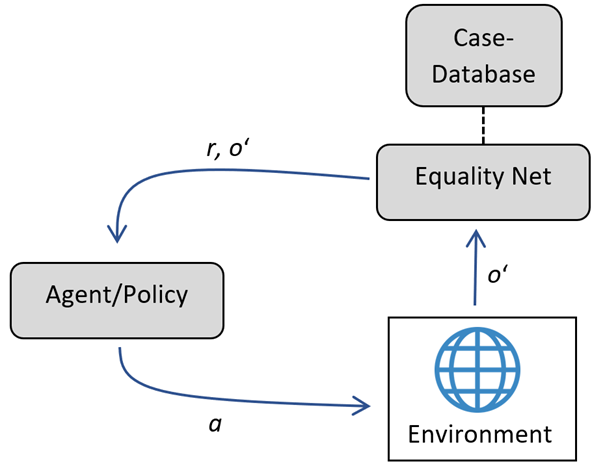}
\caption{This figure shows the usual cycle of Reinforcement Learning, with a small adjustment. The Equality Net ($E$) is interposed between the environment and the agent. The agent performs an action $a$, which is executed in the environment. $E$ receives the next observation $o'$ from the environment, then calculates the reward $r$ using the case database and forwards both to the agent.}
\end{figure}

A second advantage of Reachability Networks in contrast to embeddings is that they are suitable for asymmetric state-action spaces. For example, it may be easy to reach $s_2$ from $s_1$, but difficult or impossible to reach $s_1$ from $s_2$.
\ \\

The policy $\pi$ is learned via Inverse Reinforcement Learning using the case database $C$ and the Equality Net $E$. If the agent is in state $o$, it executes the action $a = \pi(o)$ with its current policy $\pi$ and receives the next observation $o'$ from the environment. Using $E$, all expert observations $o_e^{(i)}$ from $C$ are now checked to see if they are similar to $o'$, where the similarity must be above a threshold $\tau$. If there is a similar expert state $o_e^{(j)}$ (if more than one, choose the most similar), the reward is given by the position number $j$. Thus, the further back the similar expert state is in the expert trajectory, the higher the reward the agent receives. If there is no similar expert state, the agent receives a penalty $\mu$ (a negative reward). Figure 2 shows the idea schematically. The complete algorithm is summarized in Algorithm 1.
\ \\

The algorithm contains several hyperparameters, whose task and influence we discuss in the following: $\tau \in [0;1]$ is the threshold that determines the minimum similarity of an expert state $o_e^{(i)}$ to the current state $o$ of the agent, so that the agent receives a positive reward. If no expert state has a similarity higher than $\tau$, the agent receives a penalty (a negative reward $\mu$). The hyperparameter $\alpha \in [0;1]$ controls whether the actual reward for the agent is always the reward difference ($\alpha=1$) or whether the agent always receives the full reward ($\alpha=0$). For $\alpha=0$, the agent tends to achieve large rewards as quickly as possible, but maybe not reliably, whereas for $\alpha=1$, the agent tries to achieve a large reward as reliably as possible by the end of the episode.

The hyperparameters $\textit{windowFrame}$ and $\nu$ are used to train $E$. They model on the one hand the threshold which indicates whether two states are considered similar or dissimilar and on the other hand the number of explicit divergence between states of the agent and states of the expert.

\begin{figure}[bt]
\centering
\includegraphics[scale=0.2]{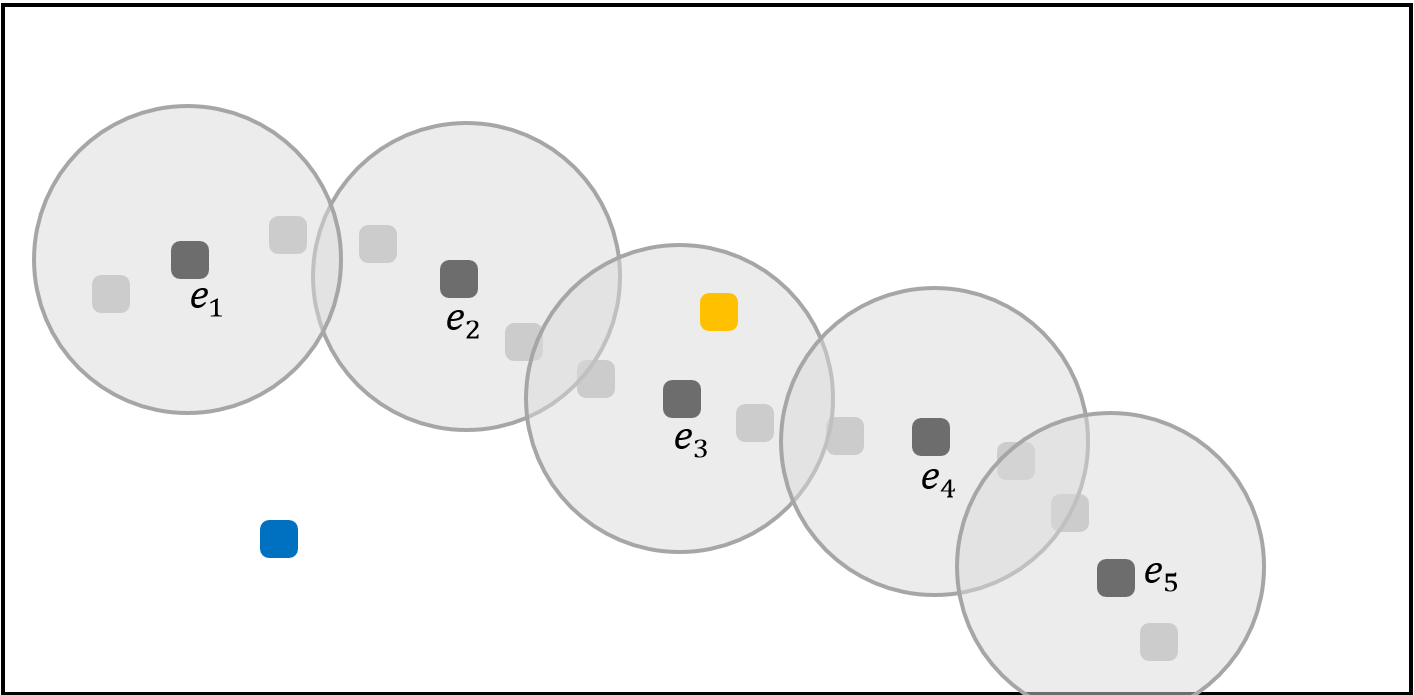}
\caption{This figure shows schematically how the algorithm works. Assume the rectangle is a two-dimensional state space. The light and dark gray boxes represent the trajectory of expert states but only the dark states are visible to the agent. Using its own rollouts the agent now learns the Equality Net, which classifies whether two states can occur close to each other in a trajectory. The outputs of the Equality Net for the expert states are represented in the image by the circles around them. During inference, the agent checks whether the current state is similar to an expert state or not. For example, if the agent is in the yellow state, it is similar to expert state $e_3$ and therefore receives reward 3. If the agent is in the blue state, it is not similar to any expert state and receives a negative reward $\mu$.}
\end{figure}
\ \\

\textbf{Training of the Equality Net}: The task of the Equality Net $E$ is to classify whether two inputs can occur in short succession in a trajectory and are thus ``similar''. To train $E$, we use the Replay Buffer that contains the trajectories sampled by the agent. $E$ is trained using supervised learning.
The training set consists of similar and dissimilar state pairs. For the similar state pairs, two states are selected from the same trajectory of the Replay Buffer which are no further apart than $\textit{windowFrame}$ steps. For the dissimilar state pairs, two states are sampled from two different trajectories. For the similar state pairs, the network $E$ is trained to output the value $1$, for dissimilar state pairs it is trained to output $0$. The structure of $E$ is graphically visualized in Figure 3.
\ \\
In addition, training can also be performed in an analogous manner on the expert trajectories. The hyperparameter $\nu$ models the number of explicit divergence between agent and expert state. That is, there are $\nu$ state pairs where one state is sampled from the Replay Buffer and the other state is sampled from $C$. The target for these state pairs during supervised learning is $0$, since they shall be classified as dissimilar.
\ \\
The output of the Equality Net can be understood as a (lossy) binary distance measure. The distance measure is binary because it only distinguishes between similar (1) and dissimilar (0).

\begin{figure}[bt]
\centering
\includegraphics[scale=0.58]{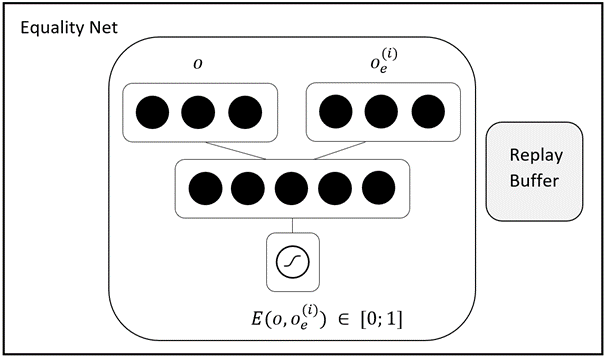}
\caption{The Equality Net $E$ accepts as input two states and classifies whether they are similar in the sense that they can appear close to each other in a trajectory. During inference, $E$ receives as input the current state $o$ of the agent and compares it to all expert states $o_e^{(i)}$. $E$ is trained using supervised learning on the trajectories produced by the agent, which are stored in the Replay Buffer.}
\end{figure}

\section{Experiments}

We tested our algorithm in four OpenAI Gym domains \cite{brockman2016openai}: Acrobot, Mountain Car, Lunar Lander, and Half Cheetah. For Half Cheetah, we created a modified version called Half Cheetah Discrete. Details can be found in Appendix A.
As justified in \cite{christiano2017deep}, only domains should be used for the evaluation of IRL algorithms in which the episodes are always of the same length. This is because early ending of episodes may contain implicit information about the reward. For example, in the `Mountain Car' domain, the episode ends when the car has successfully driven up the hill. For this reason, we have adjusted all domains so that episodes are always of the same length, with the agent receiving the last observation until the end if the episode ended early.
\ \\

We first trained an expert for each domain using the reward function of the environment. We then used these experts to create exactly one trajectory for each domain, which consisted only of the expert states and not the expert actions. We then used it to train CB-IRL and GAIFO. GAIFO had access to \underline{all} expert states, while CB-IRL only had access to \underline{every tenth} expert state. For example, for Lunar Lander, the expert trajectory was about 150 steps long, so the training set for GAIFO consisted of these 150 expert states, while the training set for CB-IRL consisted of only 15 expert states.

For the hyperparameter search, we tested five hyperparameter sets for each algorithm and domain and selected the best one. Using these hyperparameters, we then trained CB-IRL and GAIFO three times with three different seeds. During training we generated 20 episodes every 10,000 steps for each seed and algorithm (for a total of 60 episodes per algorithm every 10,000 steps). For each episode, we calculated the total return using the environment's reward function. The returns were then scaled using the performance of a random agent (representing value 0) and the expert (representing value 1). We then calculated the 0.25, 0.5 (median), and 0.75 quantiles for these 60 return values. For both algorithms, the solid lines represent the median and the shaded areas enclose the 0.25 and 0.75 quantiles.

As can be seen in Figure 4, CB-IRL mostly performed better than GAIFO in the experiments, even though it had access to only one tenth of GAIFO's training set. Furthermore, CB-IRL showed a more stable learning behavior. The difference was particularly clear in the Half Cheetah Discrete domain. Here, the advantage of CB-IRL became apparent, where the agent did not learn to behave exactly like the expert in every state, but to reach similar states as the expert. CB-IRL has learnt the high-level strategy to ``run as far as possible''.

A Python implementation of CB-IRL and the code used to create the experiments are available on GitHub [\url{https://github.com/JonasNuesslein/CB-IRL}]. For GAIFO we used the implementation of tf2rl \cite{ota2020tf2rl}. The chosen hyperparameters for the experiments can also be found on GitHub in the file $\textit{config.py}$.

\begin{figure}[bt]
\centering
\includegraphics[scale=0.42]{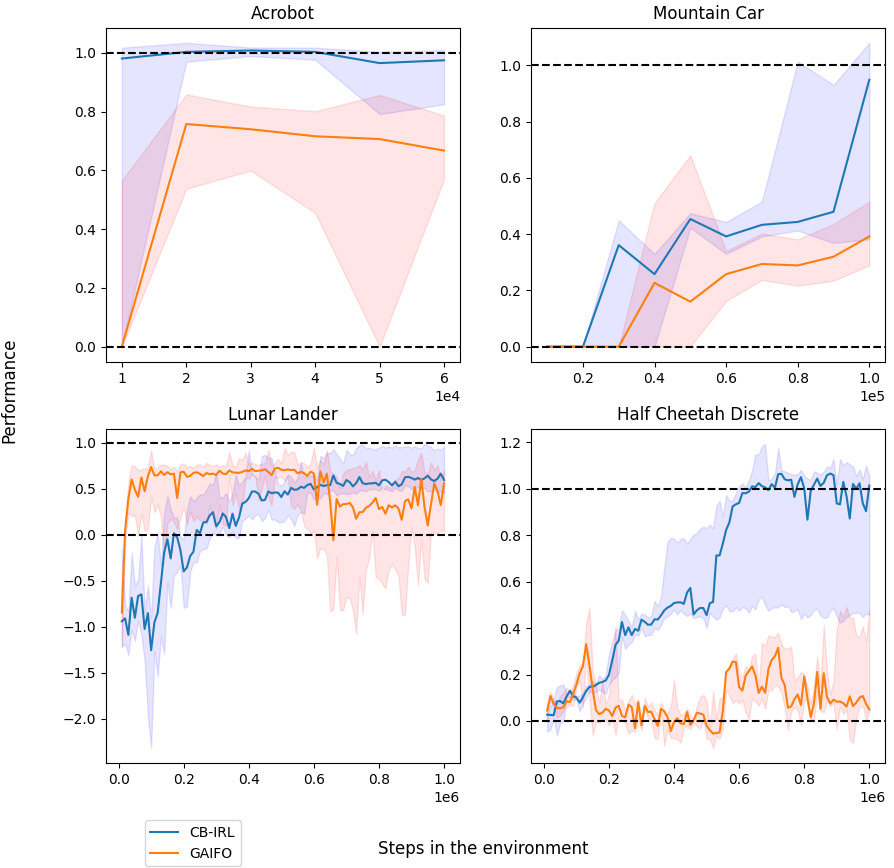}
\caption{Scaled performance of CB-IRL and GAIFO on four different domains trained using one expert trajectory, where GAIFO had access to all expert states and CB-IRL had access to only one in ten.}
\end{figure}

\section{Discussion of the Approach}

In this section we discuss the advantages and disadvantages of CB-IRL. Turning first to the disadvantages: The computation of the reward is more computationally intensive than in many other IRL algorithms, because in each step the current state must be compared against all expert states in the case base $C$. The run-time complexity is thus linear in the size of $C$. This can be serious for larger case bases, however, the target application areas of CB-IRL are precisely the settings where very little expert data is available. Moreover, the computational intensity can be reduced by calculating a reward only in every $k$-th step, rather than in every step.

The second drawback of our approach is the specialization of CB-IRL to state-reaching in contrast to state-keeping domains. By state-reaching domains, we mean domains in which certain variables of the state vector have to be changed. An example of this is the OpenAI Gym domain `Mountain Car' \cite{brockman2016openai}, in which the goal is to maximize the x-position of the car. Another example is the Atari game `Pong' \cite{mnih2015human}, in which the goal of the agent is to reach 21 points. By state-keeping domains, we mean domains in which the goal is to leave certain variables of the state vector unchanged. An example of this would be `Cart-Pole' \cite{brockman2016openai}, where the goal is to keep the angle of the pole at 90° if possible or `HalfCheetah' \cite{brockman2016openai}, where the goal is to keep a high velocity. Due to the structure of CB-IRL, it is predominantly suitable for state-reaching domains, as the algorithm encourages the agent to reach states from the posterior of the expert trajectory. \par
\ \\
The advantages of CB-IRL are that it does not require a reward function, expert actions, or complete expert trajectories.
Since the agent can learn with incomplete expert trajectories, it has proven that it imitates the higher-level strategy of the expert and does not imitate the expert on action level.

This allows the agent to learn a near-optimal strategy with little data, which would be insufficient to imitate the expert on action level (as can be seen in the Half Cheetah Discrete domain). The learning behavior also shows a more stable pattern with less ``forgetting'' of what has already been learned.

A second possible advantage, which we leave as future work to verify, is that the Equality Net is not task-specific and can be reused for other tasks in the same domain, which can enable fast transfer learning.

A third possible advantage also left for future work is that the ability to learn from incomplete trajectories may be beneficial in real-world applications, where state observations may be noisy or delayed.

\section{Conclusion}

In this paper, we have shown that when very little expert data is available, it is advantageous to imitate the higher-level strategy of the expert, rather than imitating the expert on action level.

To prove that the agent really imitated the strategy and not the expert actions, we considered a special setting of Imitation Learning characterized by incomplete expert trajectories. Moreover, no expert actions were available to the agent (Learning from Observations). The chosen prior for the higher-level strategy was to reach an unknown target state area. But since the expert has demonstrated how to reach it, the agent tries to reach similar states as the expert.

The presented algorithm Case-Based Inverse Reinforcement Learning (CB-IRL) builds on the idea of Temporal Coherence and Case-Based Reasoning. The algorithm trains a neural network to predict whether a state $s_2$ can be reached from a state $s_1$ within $\textit{windowFrame}$ time steps (actions). If so, the states can be considered ``similar''. During inference, the agent uses this network to compare its current state $o$ against expert states $o_e^{(i)}$ from a Case Base. If a similar expert state $o_e^{(j)}$ exists, the position $j$ of this expert state in the expert trajectory serves as a (positive) reward signal for the agent. If no similar expert state exists, the agent receives a penalty. Thus, the agent is trained to reach similar states to the expert states. We tested our approach on four typical domains of Reinforcement Learning, where in every case only one tenth of an expert trajectory was available to the agent. The results show that CB-IRL was able to learn a near-optimal policy, often better than GAIfO, which had access to the full expert trajectory and was trying to imitate the expert at action level.

\newpage

\bibliographystyle{alpha}
\bibliography{main}

\appendix
\section{Appendix}

The OpenAI domain Half Cheetah does not normally contain any absolute position information. To make this domain a state-reaching domain, we additionally added the x-position of the Cheetah to the otherwise 17-dimensional state space. Furthermore, the action space of this domain is originally continuous, which greatly complicates exploration. To facilitate exploration, we created a modified version called ``Half Cheetah Discrete''. For this, 20 random (continuous) action vectors were sampled from the continuous action space. These 20 action vectors can be seen as basis vectors of the original continuous action space and together they now form a discrete action space (consisting of 20 possible actions). If one of the 20 discrete actions is selected, the corresponding continuous action vector is executed in the background.

\end{document}